\documentclass[runningheads]{llncs}

\usepackage{graphicx}
\usepackage{amsmath,amssymb} 
\usepackage{hyperref}
\hypersetup{
  colorlinks=true,
  citecolor=blue
}
\usepackage{color}
\usepackage[width=122mm,left=12mm,paperwidth=146mm,height=193mm,top=12mm,paperheight=217mm]{geometry}
\begin{document}
\pagestyle{headings}
\mainmatter

\title{Speech-Driven Facial Reenactment Using Conditional  Generative Adversarial  Networks} 

\titlerunning{Speech-Driven Facial Reenactment}

\authorrunning{Jalalifar et al.}

\author{Seyed Ali Jalalifar\inst{1} \and Hosein Hasani\inst{1}
\and Hamid Aghajan\inst{1,2}}

\institute{Department of Electrical Engineering,\\Sharif University of Technology, Tehran, Iran\\
\email{\{seyedali.jalalifar,hasani.hosein\}@ee.sharif.ir}\\
\and
imec, Ghent University, Ghent, Belguim\\
\email{hamid.aghajan@ugent.be}}

\maketitle

\begin{abstract}
We present a novel approach to generating photo-realistic images of a face with accurate lip sync, given an audio input. By using a recurrent neural network, we achieved mouth landmarks based on audio features. We exploited the power of conditional generative adversarial networks to produce highly-realistic face conditioned on a set of landmarks. These two networks together are capable of producing sequence of natural faces in sync with an input audio track. 
\end{abstract}

\section{Introduction}
Creating talking heads from audio input is interesting from both scientific and practical viewpoints, e.g. constructing virtual computer generated characters, aiding hearing-impaired people, live dubbing of videos with translated audio, etc. Due to its wide variety of applications, audio to video has been the focus of intensive research in recent years \cite{Suwajanakorn,Taylor:2017:DLA:3072959.3073699,7404961,7590381}. Mapping audio to facial images with accurate lip-sync is an extremely difficult task because it is a mapping form 1-Dimensional to 3-Dimensional space and also because humans are expert at detecting any out-of-sync lip movements with respect to an audio. 

\medskip
Facial reenactment has seen considerable progress recently \cite{Thies:2016:DFR:2929464.2929475,Thies:2015:RET:2816795.2818056,GVSSVPT15,Shi:2014:AAH:2661229.2661290}. Approaches to photo-realistic facial reenactment usually involve utilizing computer graphic methods to produce high-quality results. Suwajanakorn et al.\cite{Suwajanakorn} generates photo-realistic mouth texture directly from audio using compositing techniques. In \cite{Thies:2016:DFR:2929464.2929475}, animating the facial expressions of the target video by source actor is achieved by deformation transfer between source and target. Although these methods usually produce highly-realistic reenactment, they suffer from occasional failures. A big challenge for these methods is synthesizing realistic teeth because of the subtle details in the mouth region. Unlike these approaches, we propose using pure machine learning techniques for the task of facial reenactment which we believe is more flexible and simpler to implement. By using generative adversarial networks, our model learns the manifold of human face and lip movements which is a great help for avoiding uncanny valley\footnote{Objects which closely look like humans but are different in small details elicit a sense of unfamiliarity while less similar objects look more familiar.}. 

\medskip
Generative adversarial networks (GANs), first introduced by Goodfellow et al. \cite{NIPS2014_5423}, are great tools for learning image manifold. They have shown huge potential in mimicking the underlying distribution of data, and produced visually impressive results by sampling random images drawn from the image manifold \cite{zhu2016generative,DBLP:journals/corr/abs-1710-10196,DBLP:journals/corr/MaJSSTG17,DBLP:journals/corr/ImKJM16}. Despite their power, GANs are notorious for their uncontrollable output because of the entangled space of the input data and no control on the modes of the data being generated. That was the impetus behind proposing conditional generative adversarial networks \cite{DBLP:journals/corr/MirzaO14} which offer some control over the output. Other approaches are also proposed to learn disentangled, interpretable representations in an unsupervised \cite{DBLP:journals/corr/ChenDHSSA16,DBLP:journals/corr/LarsenSW15} or supervised \cite{DBLP:journals/corr/YinFSX17} manner.

\medskip
We evaluated some extensions of generative adversarial networks and found out that conditional GAN suits best to our problem. We exploited the power of conditional GANs to generate natural faces conditioned on a set of mouth landmarks	. Another network is trained to produce mouth landmarks out of an audio input using LSTM structure. By combining these two networks, our model is capable of generating natural face with accurate lip sync. To the best of our knowledge, this is the first time that C-GANs are applied to the problem of audio to video mapping. The closest work to ours is \cite{Suwajanakorn} but unlike our method, they composited mouth texture with proper 3D pose matching for accurate lip syncing. We do a case study on a specific person, President Barak Obama, because of the huge volume of data available from his weekly address and also because the videos are online and public domain. Generating talking heads of other people is easily achievable using the same pipeline, given that enough data is available.
\section{Related Work}
The related work can be divided into two categories: Creating accurate lip sync given an audio input, and manipulating face using generative adversarial networks.
\begin{figure}[t!]
\centering
\includegraphics[height=100 px]{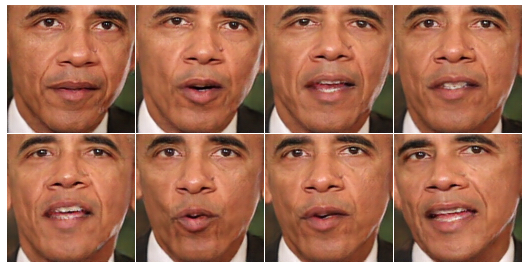}
\caption{Artificial faces of Obama, created entirly from audio input.}
\label{fig:Demo}
\end{figure}
\subsection{Lip Syncing from Audio}

Approaches to automatically generating natural looking speech animation usually involve manipulating 3D computer generated faces\cite{Liu:2015:VDR:2816795.2818122,6165277,Cao:2005:ESF:1095878.1095881}. It was not until recently that highly-realistic facial reenactment was achievable \cite{Suwajanakorn,Thies:2016:DFR:2929464.2929475}. Typical procedure to generating lip sync from audio usually consists of extracting some features from raw audio \cite{Suwajanakorn} or phoneme extraction \cite{Taylor:2017:DLA:3072959.3073699}. A mapping between audio features to 3D face model for avatars is then achieved. For the case of facial reenactment, appropriate facial texture is created from audio features. Taylor et al. \cite{Taylor:2017:DLA:3072959.3073699} proposed using a sliding window predictor that learns arbitrary non-linear mapping from phoneme label input sequence to mouth movements. Anderson et al. in \cite{Anderson:2013:ETT:2503385.2503473}, proposed a pipeline for generating text-driven 3D talking heads from limited number of 3D scans using Active Appearance Model(AAM) to construct 2D talking heads first and then create 3D models from them. One of the first highly-realistic facial reenactment approaches, Face2Face, was introduced by Thies et al. \cite{Thies:2016:DFR:2929464.2929475}. They proposed a new approach for real-time facial reenactment using monocular video sequence form source and target actor. Based on their work, Suwajanakorn et al. \cite{Suwajanakorn} introduced a new method for creating talking heads given an audio input by compositing techniques. While Face2Face transfers the mouth from another video sequence, they synthesize mouth shape directly from audio.

\medskip
Although our work is similar to \cite{Suwajanakorn} in application, there are fundamental differences between utilized methods. Conventional approaches for facial reenactment heavily involve computer graphic methods which are prone to generating uncanny faces due to the lack of understanding of the human face manifold. These methods also need to overcome some challenges related to synthesizing realistic teeth. Unlike these approaches, we propose a new pipeline for generating highly-realistic videos with accurate lip sync from audio by learning the human face manifold. This greatly reduces the complications that typical methods usually have to deal with and also prevents occasional failures.

\subsection{Generative Adversarial Networks}
Generative adversarial nets has recently received an increasing amount of attention and produced promising results, especially in the tasks of image generation \cite{zhu2016generative,DBLP:journals/corr/abs-1710-10196,DBLP:journals/corr/MaJSSTG17,DBLP:journals/corr/ImKJM16,DBLP:journals/corr/ZhangXLZHWM16} and video generation \cite{DBLP:journals/corr/TulyakovLYK17}. The power of these networks is that they produce visually impressive outputs, because they learn the underlying distribution of data. They've opened a new door to the field of image editing. Efforts for editing faces in latent space usually consist of supervised and unsupervised methods to disentangle the latent space. Chen et al. \cite{DBLP:journals/corr/ChenDHSSA16} proposed an information-theoretic extension of GANs, InfoGAN, which is able to learn disentangled representation of latent space in a completely unsupervised manner. This is done by maximizing mutual information between some latent variables and the observation. They tested their approach on the CelebA dataset and managed to control pose, presence or absence of glasses, hair style and emotion of generated face images. Semi-Latent GAN, proposed by Yin et al. \cite{DBLP:journals/corr/YinFSX17}, learns to generate and modify images from attributes by decomposing noise of GAN into two parts, user defined attributes and latent attributes, which are obtained from the data. 

\medskip
GAN-based conditional image generation has also been the focus of research in recent years. In conditional GANs, both generator and discriminator are provided with class information. Ma et al. in \cite{DBLP:journals/corr/MaJSSTG17} proposed a method for pose guided person image generation conditioned on a specific pose. Kaneko et al. \cite{8100224} presented a generative attribute controller by utilizing conditional filtered generative adversarial networks. In this paper, we use conditional GANs to generate facial images, given a set of landmarks. Our model is capable of generating faces with accurate alignment with given landmarks. Another LSTM network learns to predict facial landmark positions from audio features. These two networks together are able to generate realistic facial images with accurate lip sync, given an audio input.
\begin{figure}[t!]
\centering
\includegraphics[height=3cm]{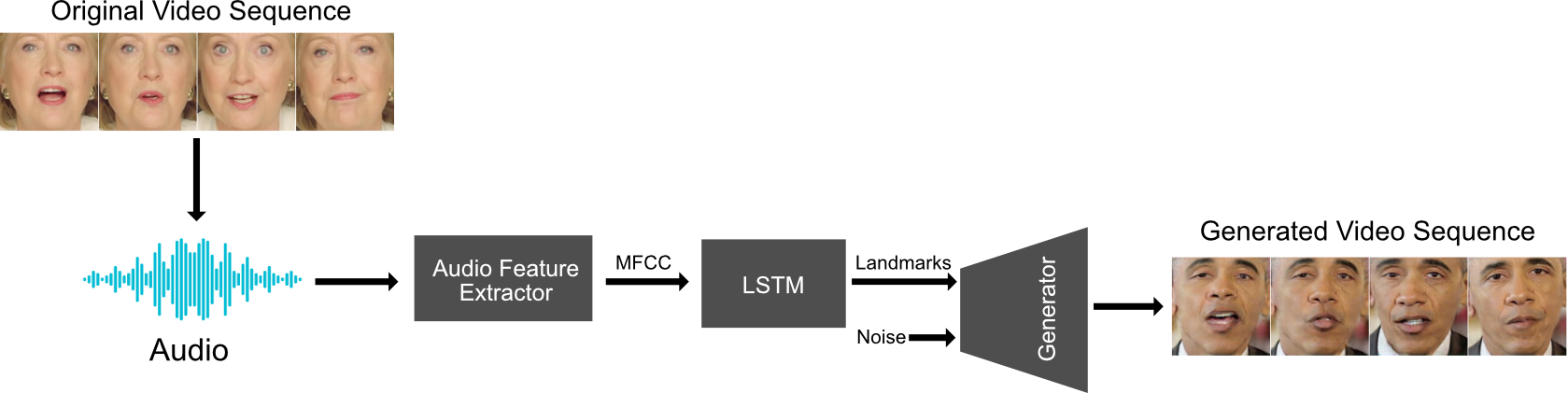}
\caption{An overview of the proposed system. First an LSTM network is trained with audio features as input and lip landmark positions as labels. A C-GAN is trained to produce highly-realistic faces with respect to a given set of landmarks. Finally, these two networks together are able to produce convincing faces from an audio track.}
\label{fig:System_Overview}
\end{figure}
\section{System Overview}
\label{sec:overview}

An overview of the system is shown at Fig. 2. At the heart of our system is a conditional GAN which is trained to produce highly realistic facial images conditioned on a given set of lip landmarks. An LSTM network is utilized to create lip landmarks out of audio input. Here we briefly introduce the implemented networks  in our system.

\subsection {LSTM}
Long Short-Term Memory networks, first introduced by Hochreiter et al. \cite{Hochreiter:1997:LSM:1246443.1246450}, are a special type of recurrent neural networks. Unlike typical networks, RNNs have the ability to connect previous information to the present task. While RNNs are not capable of handling long-term dependencies \cite{Bengio:1994:LLD:2325857.2328340}, LSTMs are explicitly designed to handle such situations. The computation within an LSTM cell can be described as:
\begin{gather}
f_{t+1} = \sigma{(W_f.[h_t,\Phi_t]+b_f)}, \\
i_{t+1} = \sigma{(W_i.[h_t,\Phi_t]+b_i)}, \\
o_{t+1} = \sigma{(W_o.[h_t,\Phi_t]+b_o)}, \\
\tilde{C}_{t+1} = tanh{(W_c.[h_t,\Phi_t]+b_c)}.
  \end{gather}
where $C_t$, $h_t$ and $\Phi_t$ are the inputs to the LSTM.  $W_f$,$W_i$,$W_o$,$b_i$,$b_o$,$b_f$, and $b_c$ are trainable  parameters. $\sigma$ is the sigmoid activation function. $f,i,o$ are the forgetting, input and output gates of an standard LSTM unit which control the contribution of historical information to current decision. The outputs of an LSTM cell are 
\begin{gather}
C_{t+1} =  f_{t+1}C_t+i_{t+1} \tilde{C}_{t+1},\\
h_{t+1} = o_{t+1} tanh(C_{t+1}). 
\end{gather}

\subsection{Conditional Generative Adversarial Networks}
Typically, A GAN consists of two networks, a generator and a discriminator. The generator network G tries to fool the discriminator D by creating samples as if they come from the real distribution of data. It is discriminator's task to distinguish between fake and real samples while the generator tries to learn the true distribution of data in order to fool the discriminator. As training goes on, the discriminator becomes better and better at dividing real and fake samples so the generator has to produce more realistic samples to deceive the discriminator. This leads to a two player min-max game with the value function $V(G,D)$:
\begin{gather}
\min_{G}\max_{D}V(G,D) = \mathbb{E}_{x\sim{p_{data}(x)}}[log(D(x)]+\mathbb{E}_{z\sim{p_{z}(z)}}[log(1-D(G(z)))]. 
\end{gather}

GANs find a mapping between prior noise distribution to data space. Since values of latent code are picked randomly from a distribution, there is no control over the output of generator. Conditioning both discriminator and generator on some extra information $y$ offers some control over the output \cite{DBLP:journals/corr/MirzaO14}. $y$ could be any kind of auxiliary information, such as class label or as in our case, landmark positions. In the case of conditional GANs, the objective function of a two-player min-max game would be:
\begin{gather}
\min_{G}\max_{D}V(G,D) = \mathbb{E}_{x\sim{p_{data}(x)}}[log(D(x|y)]+\mathbb{E}_{z\sim{p_{z}(z)}}[log(1-D(G(z|y)))]. 
\end{gather}
After training GAN network, the discriminator network is discarded and only the generator is used for creating realistic facial images.
\section{Approach}
Mapping a sequence of audio to a sequence of images is inherently a difficult task due to the ambiguities of mapping from low-dimensional to high-dimensional space. Our ultimate goal is to estimate the distribution $p_{model}(x_i|V_i)$ where $x_i$ is an image at $i^{th}$ frame, and $V_i=[v_{i-n},v_{i-n+1},...,v_{i+n}]$ is audio feature vector with $2n+1$ as sequence size. Instead of directly computing $p_{model}$, we try to estimate distributions $p_\theta(l_i|V_i)$ and $p_\phi(x_i|l_i)$ where $l_i$ consists of 8 landmark positions. Now the problem is finding $\theta$ and $\phi$ which represent model parameters of LSTM and Generator networks respectively. First an LSTM network is trained to output facial landmark positions based on the mel-frequency cepstral coefficients extracted from audio input. Another generative model is trained to create high-quality realistic faces conditioned on a set of landmarks. These two networks are trained independently and result in a mapping from MFCC audio features to a sequence of facial images in sync with a given audio.

\subsection{Data Acquisition}
For the training part, we used President Obama's weekly address videos because of their availability, high quality and controlled environment. These videos are 14 hours in total but we used a subset of the dataset since we achieved the desired quality with about two hours of videos. For each frame, we extracted the face region, in addition to important lip landmarks with the method proposed in \cite{Kazemi:2014:OMF:2679600.2679766}.We also extracted  mel-frequency cepstral coefficients from audio input.

\subsection{Audio to Landmarks}
 The shape of the mouth during speech depends not only on the current phoneme but also on the phonemes before and after. This is called co-articulation and it can affect up to 10 neighboring phonemes. Inspired by \cite{7404961} and \cite{Suwajanakorn}, we used LSTM for preserving these long-term dependencies. Extensive work has been done on the problem of audio feature extraction \cite{Lee:2009:UFL:2984093.2984217,10.1007/978-3-319-11581-8_26,DBLP:journals/corr/TakahashiGG17}. We used the typical mel-frequency cepstral coefficients as the audio features. We took discrete Fourier transform on every 33 milliseconds sliding window and applied 40 triangular mel-scale filters to the Fourier power spectrum. In addition to these 13 MFCCs, we also used their first temporal deviation and log mean energy as extra features to obtain a 28-D feature vector.
From the 68 landmark points detected by Dlib \cite{King:2009:DML:1577069.1755843}, we selected the most correlated ones with speech which are 8 points around the lip. These 8 points make up a 16-D vector. We used a single layer LSTM structure followed by two hidden layers for mapping from audio to the lip landmarks. More details about the implementation can be found on Section 5.

\subsection{Landmarks to Image}
We propose using conditional generative adversarial networks to create image from landmarks. We used the position of distinctive lip landmarks as an extra condition on the generator network. The input of generator consists of a 50-D noise vector and a 16-D vector of landmark positions. The 66-D input vector ultimately becomes a 128x128x3 image through deconvolution networks. We concatenate the resultant image with 16-D landmark positions in a way that the input shape of discriminator finally becomes 128x128x19. We followed the typical network structure proposed for DC-GANs except that we concatenated both generator and discriminator network inputs with landmark positions. The structure of generator and discriminator network is shown at Fig. 3.

\begin{figure}[t!]
\centering
\includegraphics[height=6cm]{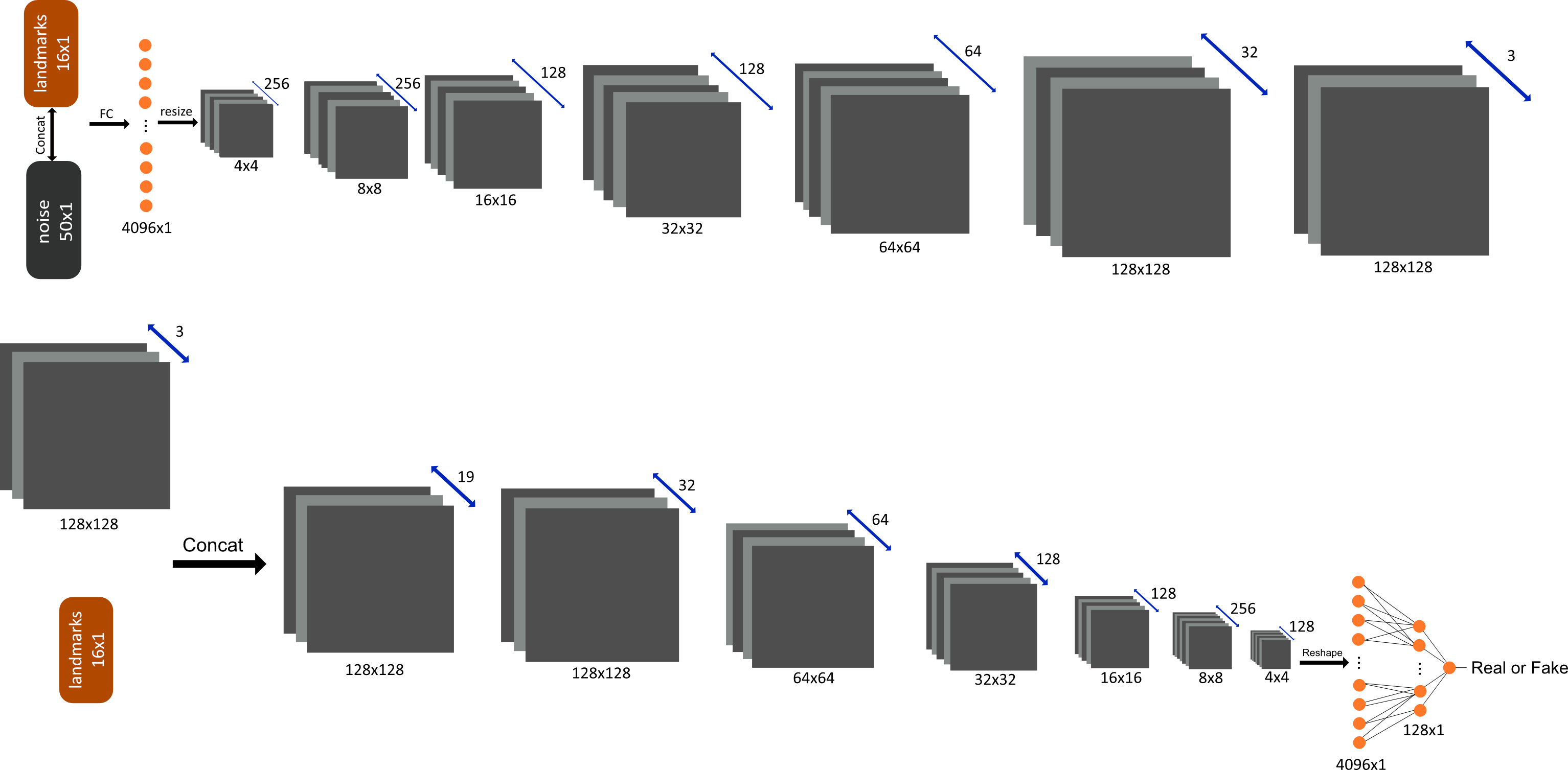}
\caption{Our conditional GAN network overview. Deconvolutional network of Generator (Top), Convolutional network of Discriminator (Bottom).}
\label{fig:example}
\end{figure}

\section{Results}
In this part we discuss implementation details and results.
\begin{figure}
\centering
\includegraphics[height=3cm]{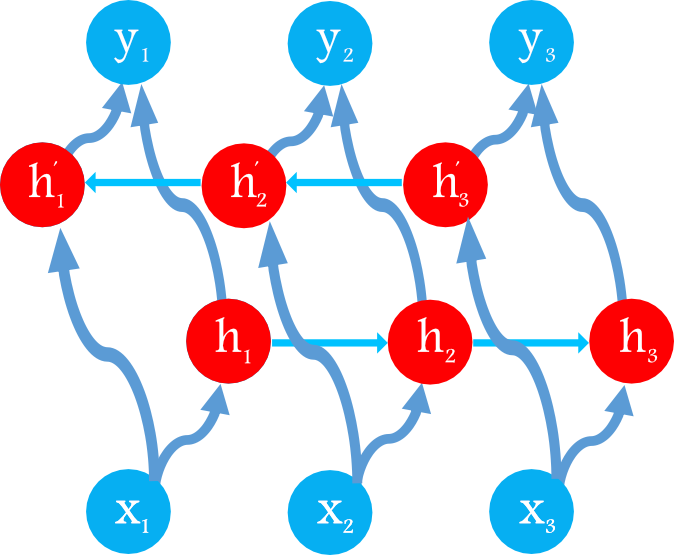}
\caption{Bidirectional LSTM structure. Frames after and before the current frame effect the output.}
\label{fig:example}
\end{figure}
\subsection{Implementation of LSTM}
We tested some architecture of LSTM and find out that bidirectional LSTM suits best to our problem. As mentioned in Section 4.2, the phenomenon called co-articulation causes mouth shape to be dependent on phonemes before and after the current phoneme. The choice of bidirectional LSTM is rational since it takes into account previous and next frames. We used a single layer bidirectional LSTM since it produces the desired quality and there is no need to introduce complexity to the network by adding extra layers. We used Adam optimizer \cite{DBLP:journals/corr/KingmaB14} for training using Tensorflow framework \cite{tensorflow2015-whitepaper}. Fig. 4 shows the bidirectional LSTM structure. In table 1, we compare performance of different LSTM structures and parameters.
\begin{table}
\caption{Validation loss of different LSTM network structures}
\centering
\begin{tabular}{lrrr}
\hline
\multicolumn{4}{c}{}{Validation Loss(Epochs)} \\
\cline{2-4}
Network Structure   & 100 epochs& 200 epochs&300 epochs \\
\hline
Single-layer bidirectional LSTM    & 0.91    & 0.88   & 0.85   \\
Single-layer unidirectional LSTM   & 0.93    & 0.91   & 0.93    \\
Two-layer bidirectional LSTM       & 0.92    & 0.88   & 0.84     \\
\hline
\end{tabular}
\end{table}

\begin{table}
\caption{Validation loss of different LSTM networks versus dropout probability}
\def\arraystretch{1.5}
\centering
    \begin{tabular}{| c | c | c | c |}
    \hline
    Dropout rate: & 0 & 0.3 & 0.5 \\ \hline
    Single-layer bidirectional LSTM & 0.91 & 0.88 & 0.93\\ \hline
    Single-layer unidirectional LSTM & 0.94 & 0.92 &0.95 \\ \hline
    Two-layer bidirectional LSTM & 0.91 & 0.89 & 0.92 \\
    \hline
    \end{tabular}
\end{table}

\subsection{Conditional Generative Adversarial Network}

Our conditional network is able to create real facial images out of landmarks. While generating image sequence from audio input, we need to keep facial texture and background constant. In order to achieve so, we limit C-GAN training dataset in the last epochs to the target video. This keeps the facial texture constant during face generation while preserves the details of the reconstructed face. Some tricks proposed in \cite{odena2016deconvolution} improved the quality of the output and reduced visual artifacts. Fig. 5 shows some of the results we achieved from a given set of landmarks.

\medskip
The novelty of our approach is that the two modules that we used, LSTM and C-GAN, are almost independent from each other. This means that our model is able to transfer lip movements of other people, given their audio. Only a simple affine transformation should be applied to the source facial landmarks in order to be aligned with the target landmarks. Fig. 6 shows transfer from Hillary Clinton’s audio speech to President Barack Obama’s lip movements.
\begin{figure}[t!]
\centering
\includegraphics[height=100 px]{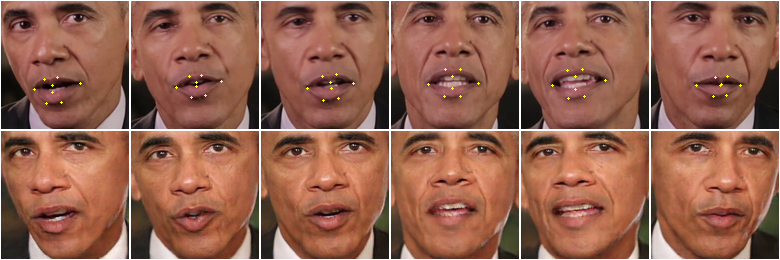}
\caption{Images directly generated from landmarks. Original sequence (Top), Generated face using C-GAN (Buttom).}
\label{fig:example}
\end{figure}
\begin{figure}[t!]
\centering
\includegraphics[height=140 px]{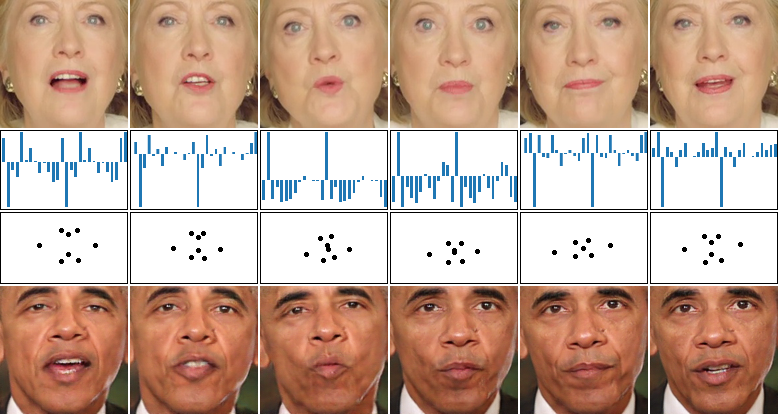}
\caption{Creating Artificial faces of President Obama, given an audio track from Hillary Clinton. From top to bottom: 1) Original video, 2) Audio features, 3) Predicted landmarks, 4) Generated Images created from landmarks.}
\label{fig:example}
\end{figure}

\section{Conclusion, Limitations and Future Work}

We propose using conditional generative adversarial networks for creating high-quality faces given their mouth landmarks. The mouth landmarks are also obtained from audio using an LSTM network. This gives us an end-to-end system with much flexibility, e.g. the ability to manipulate faces without losing their naturalness. This is a huge advantage over computer graphic methods since there is no need to get involved with details of face, e.g. synthesizing realistic teeth. The LSTM network and C-GAN network are almost independent from each other so we can reanimate target face with audios from other sources rather than the target person himself. This opens the door for many interesting new applications such as face transformation, Dubsmash like apps, etc.
\medskip

We used Dlib landmark detector for extracting facial landmarks. There are new approaches with more accurate results available for facial landmark detection, especially in the mouth region \cite{Xiong-2013-7701}. Using these improved methods increase the quality of the LSTM network to predict mouth shape from audio features.
\medskip

Sometimes our model fails to create natural faces. This is mainly because the fact that the provided lip landmarks are significantly different from what the C-GAN saw during training phase (Fig. 7). To address this problem, a more comprehensive dataset can be used to cover more head poses and lip landmark positions.
\medskip

Finally, typical DC-GAN structure and training procedure are utilized during training phase. New architectures and algorithms such as \cite{DBLP:journals/corr/abs-1710-10196} has been proposed to improve the quality of output image. Using these new structures, images with higher quality and finer details are achievable.
\begin{figure}
\centering
\includegraphics[height=64 px]{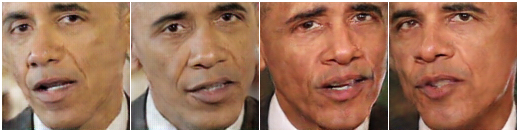}
\caption{Some cases of failiure mainly caused by irrelevent lip landmarks.}
\label{fig:example}
\end{figure}

\nocite{*}
\bibliographystyle{splncs}
\bibliography{egbib}

\begin{thebibliography}{10}

\bibitem{Suwajanakorn}
Suwajanakorn, S., Seitz, S.M., Kemelmacher-Shlizerman, I.:
\newblock Synthesizing obama: Learning lip sync from audio.
\newblock ACM Trans. Graph. \textbf{36}(4) (July 2017)  95:1--95:13

\bibitem{Taylor:2017:DLA:3072959.3073699}
Taylor, S., Kim, T., Yue, Y., Mahler, M., Krahe, J., Rodriguez, A.G., Hodgins,
  J., Matthews, I.:
\newblock A deep learning approach for generalized speech animation.
\newblock ACM Trans. Graph. \textbf{36}(4) (July 2017)  93:1--93:11

\bibitem{7404961}
Shimba, T., Sakurai, R., Yamazoe, H., Lee, J.H.:
\newblock Talking heads synthesis from audio with deep neural networks.
\newblock In: 2015 IEEE/SICE International Symposium on System Integration
  (SII). (Dec 2015)  100--105

\bibitem{7590381}
Llorach, G., Evans, A., Blat, J., Grimm, G., Hohmann, V.:
\newblock Web-based live speech-driven lip-sync.
\newblock In: 2016 8th International Conference on Games and Virtual Worlds for
  Serious Applications (VS-GAMES). (Sept 2016)  1--4

\bibitem{Thies:2016:DFR:2929464.2929475}
Thies, J., Zollhfer, M., Stamminger, M., Theobalt, C., Niessner, M.:
\newblock Demo of face2face: Real-time face capture and reenactment of rgb
  videos.
\newblock In: ACM SIGGRAPH 2016 Emerging Technologies. SIGGRAPH '16, New York,
  NY, USA, ACM (2016)  5:1--5:2

\bibitem{Thies:2015:RET:2816795.2818056}
Thies, J., Zollhfer, M., Niessner, M., Valgaerts, L., Stamminger, M., Theobalt,
  C.:
\newblock Real-time expression transfer for facial reenactment.
\newblock ACM Trans. Graph. \textbf{34}(6) (October 2015)  183:1--183:14

\bibitem{GVSSVPT15}
Garrido, P., Valgaerts, L., Sarmadi, H., Steiner, I., Varanasi, K., Perez, P.,
  Theobalt, C.:
\newblock Vdub: Modifying face video of actors for plausible visual alignment
  to a dubbed audio track.
\newblock \textbf{34}(2) (2015)  193--204

\bibitem{Shi:2014:AAH:2661229.2661290}
Shi, F., Wu, H.T., Tong, X., Chai, J.:
\newblock Automatic acquisition of high-fidelity facial performances using
  monocular videos.
\newblock ACM Trans. Graph. \textbf{33}(6) (November 2014)  222:1--222:13

\bibitem{NIPS2014_5423}
Goodfellow, I., Pouget-Abadie, J., Mirza, M., Xu, B., Warde-Farley, D., Ozair,
  S., Courville, A., Bengio, Y.:
\newblock Generative adversarial nets.
\newblock In Ghahramani, Z., Welling, M., Cortes, C., Lawrence, N.D.,
  Weinberger, K.Q., eds.: Advances in Neural Information Processing Systems 27.
\newblock Curran Associates, Inc. (2014)  2672--2680

\bibitem{zhu2016generative}
Zhu, J.Y., Kr{\"a}henb{\"u}hl, P., Shechtman, E., Efros, A.A.:
\newblock Generative visual manipulation on the natural image manifold.
\newblock In: Proceedings of European Conference on Computer Vision (ECCV).
  (2016)

\bibitem{DBLP:journals/corr/abs-1710-10196}
Karras, T., Aila, T., Laine, S., Lehtinen, J.:
\newblock Progressive growing of gans for improved quality, stability, and
  variation.
\newblock CoRR \textbf{abs/1710.10196} (2017)

\bibitem{DBLP:journals/corr/MaJSSTG17}
Ma, L., Jia, X., Sun, Q., Schiele, B., Tuytelaars, T., Gool, L.V.:
\newblock Pose guided person image generation.
\newblock CoRR \textbf{abs/1705.09368} (2017)

\bibitem{DBLP:journals/corr/ImKJM16}
Im, D.J., Kim, C.D., Jiang, H., Memisevic, R.:
\newblock Generating images with recurrent adversarial networks.
\newblock CoRR \textbf{abs/1602.05110} (2016)

\bibitem{DBLP:journals/corr/MirzaO14}
Mirza, M., Osindero, S.:
\newblock Conditional generative adversarial nets.
\newblock CoRR \textbf{abs/1411.1784} (2014)

\bibitem{DBLP:journals/corr/ChenDHSSA16}
Chen, X., Duan, Y., Houthooft, R., Schulman, J., Sutskever, I., Abbeel, P.:
\newblock Infogan: Interpretable representation learning by information
  maximizing generative adversarial nets.
\newblock CoRR \textbf{abs/1606.03657} (2016)

\bibitem{DBLP:journals/corr/LarsenSW15}
Larsen, A.B.L., S{\o}nderby, S.K., Winther, O.:
\newblock Autoencoding beyond pixels using a learned similarity metric.
\newblock CoRR \textbf{abs/1512.09300} (2015)

\bibitem{DBLP:journals/corr/YinFSX17}
Yin, W., Fu, Y., Sigal, L., Xue, X.:
\newblock Semi-latent {GAN:} learning to generate and modify facial images from
  attributes.
\newblock CoRR \textbf{abs/1704.02166} (2017)

\bibitem{Liu:2015:VDR:2816795.2818122}
Liu, Y., Xu, F., Chai, J., Tong, X., Wang, L., Huo, Q.:
\newblock Video-audio driven real-time facial animation.
\newblock ACM Trans. Graph. \textbf{34}(6) (October 2015)  182:1--182:10

\bibitem{6165277}
Le, B.H., Ma, X., Deng, Z.:
\newblock Live speech driven head-and-eye motion generators.
\newblock IEEE Transactions on Visualization and Computer Graphics
  \textbf{18}(11) (Nov 2012)  1902--1914

\bibitem{Cao:2005:ESF:1095878.1095881}
Cao, Y., Tien, W.C., Faloutsos, P., Pighin, F.:
\newblock Expressive speech-driven facial animation.
\newblock ACM Trans. Graph. \textbf{24}(4) (October 2005)  1283--1302

\bibitem{Anderson:2013:ETT:2503385.2503473}
Anderson, R., Stenger, B., Wan, V., Cipolla, R.:
\newblock An expressive text-driven 3d talking head.
\newblock In: ACM SIGGRAPH 2013 Posters. SIGGRAPH '13, New York, NY, USA, ACM
  (2013)  80:1--80:1

\bibitem{DBLP:journals/corr/ZhangXLZHWM16}
Zhang, H., Xu, T., Li, H., Zhang, S., Huang, X., Wang, X., Metaxas, D.N.:
\newblock Stackgan: Text to photo-realistic image synthesis with stacked
  generative adversarial networks.
\newblock CoRR \textbf{abs/1612.03242} (2016)

\bibitem{DBLP:journals/corr/TulyakovLYK17}
Tulyakov, S., Liu, M., Yang, X., Kautz, J.:
\newblock Mocogan: Decomposing motion and content for video generation.
\newblock CoRR \textbf{abs/1707.04993} (2017)

\bibitem{8100224}
Kaneko, T., Hiramatsu, K., Kashino, K.:
\newblock Generative attribute controller with conditional filtered generative
  adversarial networks.
\newblock In: 2017 IEEE Conference on Computer Vision and Pattern Recognition
  (CVPR). (July 2017)  7006--7015

\bibitem{Hochreiter:1997:LSM:1246443.1246450}
Hochreiter, S., Schmidhuber, J.:
\newblock Long short-term memory.
\newblock Neural Comput. \textbf{9}(8) (November 1997)  1735--1780

\bibitem{Bengio:1994:LLD:2325857.2328340}
Bengio, Y., Simard, P., Frasconi, P.:
\newblock Learning long-term dependencies with gradient descent is difficult.
\newblock Trans. Neur. Netw. \textbf{5}(2) (March 1994)  157--166

\bibitem{Kazemi:2014:OMF:2679600.2679766}
Kazemi, V., Sullivan, J.:
\newblock One millisecond face alignment with an ensemble of regression trees.
\newblock In: Proceedings of the 2014 IEEE Conference on Computer Vision and
  Pattern Recognition. CVPR '14, Washington, DC, USA, IEEE Computer Society
  (2014)  1867--1874

\bibitem{Lee:2009:UFL:2984093.2984217}
Lee, H., Largman, Y., Pham, P., Ng, A.Y.:
\newblock Unsupervised feature learning for audio classification using
  convolutional deep belief networks.
\newblock In: Proceedings of the 22Nd International Conference on Neural
  Information Processing Systems. NIPS'09, USA, Curran Associates Inc. (2009)
  1096--1104

\bibitem{10.1007/978-3-319-11581-8_26}
Pale{\v{c}}ek, K.:
\newblock Extraction of features for lip-reading using autoencoders.
\newblock In Ronzhin, A., Potapova, R., Delic, V., eds.: Speech and Computer,
  Cham, Springer International Publishing (2014)  209--216

\bibitem{DBLP:journals/corr/TakahashiGG17}
Takahashi, N., Gygli, M., Gool, L.V.:
\newblock Aenet: Learning deep audio features for video analysis.
\newblock CoRR \textbf{abs/1701.00599} (2017)

\bibitem{King:2009:DML:1577069.1755843}
King, D.E.:
\newblock Dlib-ml: A machine learning toolkit.
\newblock J. Mach. Learn. Res. \textbf{10} (December 2009)  1755--1758

\bibitem{DBLP:journals/corr/KingmaB14}
Kingma, D.P., Ba, J.:
\newblock Adam: {A} method for stochastic optimization.
\newblock CoRR \textbf{abs/1412.6980} (2014)

\bibitem{tensorflow2015-whitepaper}
Abadi, M., Agarwal, A., Barham, P., Brevdo, E., Chen, Z., Citro, C., Corrado,
  G.S., Davis, A., Dean, J., Devin, M., Ghemawat, S., Goodfellow, I., Harp, A.,
  Irving, G., Isard, M., Jia, Y., Jozefowicz, R., Kaiser, L., Kudlur, M.,
  Levenberg, J., Man\'{e}, D., Monga, R., Moore, S., Murray, D., Olah, C.,
  Schuster, M., Shlens, J., Steiner, B., Sutskever, I., Talwar, K., Tucker, P.,
  Vanhoucke, V., Vasudevan, V., Vi\'{e}gas, F., Vinyals, O., Warden, P.,
  Wattenberg, M., Wicke, M., Yu, Y., Zheng, X.:
\newblock {TensorFlow}: Large-scale machine learning on heterogeneous systems
  (2015) Software available from tensorflow.org.

\bibitem{odena2016deconvolution}
Odena, A., Dumoulin, V., Olah, C.:
\newblock Deconvolution and checkerboard artifacts.
\newblock Distill (2016)

\bibitem{Xiong-2013-7701}
Xiong, X., la~Torre~Frade, F.D.:
\newblock Supervised descent method and its applications to face alignment.
\newblock In: IEEE International Conference on Computer Vision and Pattern
  Recognition (CVPR). (May 2013)

\bibitem{4156221}
Umapathy, K., Krishnan, S., Rao, R.K.:
\newblock Audio signal feature extraction and classification using local
  discriminant bases.
\newblock IEEE Transactions on Audio, Speech, and Language Processing
  \textbf{15}(4) (May 2007)  1236--1246

\end{thebibliography}

\end{document}